\pdfoutput=1
\documentclass[graybox]{svmult}

\usepackage{mathptmx}       % selects Times Roman as basic font
\usepackage{helvet}         % selects Helvetica as sans-serif font
\usepackage{courier}        % selects Courier as typewriter font
\usepackage{type1cm}        % activate if the above 3 fonts are
\usepackage{makeidx}         % allows index generation
\usepackage{graphicx}        % standard LaTeX graphics tool
\usepackage{multicol}        % used for the two-column index
\usepackage[bottom]{footmisc}% places footnotes at page bottom
\usepackage{avm}
\usepackage{cite}
\usepackage{textcomp}

\makeindex             % used for the subject index
\begin{document}

\title*{FastV2C-HandNet : Fast Voxel to Coordinate Hand Pose Estimation with 3D Convolutional Neural Networks}
\titlerunning{FastV2C-HandNet}
\author{Rohan Lekhwani, Bhupendra Singh}
\institute{Rohan Lekhwani \at Indian Institute of Information Technology Pune \email{rohanlekhwani@gmail.com}
\and Bhupendra Singh \at Indian Institute of Information Technology Pune \email{bhupi.pal08@gmail.com}}
\maketitle

\abstract*{Hand pose estimation from monocular depth images has been an important and challenging problem in the Computer Vision community. In this paper, we present a novel approach to estimate 3D hand joint locations from 2D depth images. Unlike most of the previous methods, our model captures the 3D spatial information from a depth image using 3D CNNs thereby giving it a greater understanding of the input. We voxelize the input depth map to capture the 3D features of the input and perform 3D data augmentations to make our network robust to real-world images. Our network is trained in an end-to-end manner which reduces time and space complexity significantly when compared to other methods. Through extensive experiments, we show that our model outperforms state-of-the-art methods with respect to the time it takes to train and predict 3D hand joint locations. This makes our method more suitable for real-world hand pose estimation scenarios.}

\abstract{Hand pose estimation from monocular depth images has been an important and challenging problem in the Computer Vision community. In this paper, we present a novel approach to estimate 3D hand joint locations from 2D depth images. Unlike most of the previous methods, our model captures the 3D spatial information from a depth image using 3D CNNs thereby giving it a greater understanding of the input. We voxelize the input depth map to capture the 3D features of the input and perform 3D data augmentations to make our network robust to real-world images. Our network is trained in an end-to-end manner which reduces time and space complexity significantly when compared to other methods. Through extensive experiments, we show that our model outperforms state-of-the-art methods with respect to the time it takes to train and predict 3D hand joint locations. This makes our method more suitable for real-world hand pose estimation scenarios.}

\section{Introduction}
\label{intro}
Accurate human hand pose estimation forms an important task in Human Computer Interaction (HCI) and Augmented Reality (AR) systems and is one of the most highlighted problems of the Computer Vision community that is garnering a lot of attention recently \cite{Ref4, Ref5, Ref6, Ref10, Ref12, Ref18, Ref25, Ref27, Ref28, Ref41, Ref43}. Hand pose estimation forms the core of several interesting applications such as sign-language recognition \cite{Ref22, Ref29}, aerial keyboards \cite{Ref30, Ref35} and driver hand-gesture analyses \cite{Ref37, Ref39}. 

With the advent of Convolutional Neural Networks (CNNs) to be used for object classification \cite{Ref1}, there has been a considerable boost in depth-based hand pose estimation methods using 2D CNNs \cite{Ref3, Ref5, Ref13, Ref28, Ref41, Ref44}. Most of these methods take input as a 2D depth image and regress 3D joint coordinates directly \cite{Ref4, Ref16, Ref27, Ref28, Ref38}. The main problem with methods using 2D CNNs for hand pose estimation is that they try to map a 2D depth image to regress 3D world coordinates directly. This causes significant loss of information as the depth images intrinsically depict 3D coordinates of the hand with respect to the depth camera and interpreting them in two-dimensions misses out on important spatial information. Methods \cite{Ref11, Ref43} use multi-view CNNs and a feedback-loop respectively to account for the lost information but this makes it a multistage approach which increases the time and compute taken for results. Most of these methods are limited to the dataset they are trained on and hence do not generalize well on real-world scenarios.

\begin{figure}[t]
\sidecaption[t]
% For example, with the graphicx style use
\includegraphics[scale=.5]{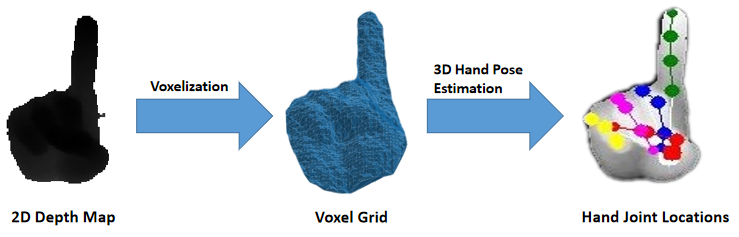}
%
% If no graphics program available, insert a blank space i.e. use
%\picplace{5cm}{2cm} % Give the correct figure height and width in cm
%
\caption{Overview of our proposed method. The 2D hand depth image after localization is projected into 3D world coordinates which are then voxelized as described in Section \ref{input}. This voxelized grid is then fed to our network after data augmentations which regresses joint locations. The ground truth joint locations on the depth image are shown in the above figure for the purpose of representation.}
\label{fig:1}       % Give a unique label
\end{figure}

In this paper, we propose a novel voxel-to-coordinate approach that can efficiently capture the spatial information of the depth image. An overview of the proposed method is shown in Figure \ref{fig:1}.The proposed method is end-to-end and reduces the time and compute power significantly opposed to previously used methods, as shown through the experiments performed in Section \ref{exp}. Voxelizing the input depth map retains the 3D nature of the input and provides the network with greater information as shown by \cite{Ref6}. Specifically, we segment the hand in the depth map from its surroundings using depth-thresholding and use reference points to localize the hand. We design a deep network to capture the 3D essence of the depth image and predict accurate joint locations. The depth image of the hand is cropped and fed to this network after performing 3D data augmentations to make our network robust to real-world scenarios. The network regresses 3D joint coordinates faster than previous methods. We evaluate our method on a publicly available benchmark dataset \cite{Ref19} and through experiments quantify the reduction in complexity by our network.

Our contributions can be summarized as follows:
\begin{enumerate}
\item{We propose a voxel-to-coordinate architecture(FastV2C - HandNet) which efficiently captures 3D information as opposed to 2D CNN methods and efficiently regresses 3D world joint coordinates. We convert the input depth map into a voxel grid which is then fed to the network.}
\item{We empirically show that our method reduces the time complexity and compute power significantly when compared to methods. This makes possible faster achievement of results while retaining the whole information of the input.}
\item{We evaluate our method on a publicly available benchmark dataset \cite{Ref19}. We also perform three-dimensional data augmentations to better generalize our model and make it more robust when used in real-time applications.}
\end{enumerate}

The remainder of the paper is organized as follows:
Previous related work has been discussed in Section \ref{relatedwork}. Section \ref{input} elaborates upon the steps used to localize the hand, voxelize it and perform 3D data augmentations. The network architecture of our model along with its implementation is discussed in Section \ref{method}. Experiments in Section \ref{exp} provides experimental results and Section \ref{comp} compares the proposed model with state-of-the-art methods. Section \ref{conc} concludes the paper.

\section{Related Work}
\label{relatedwork}
There is a significant work done on hand pose estimation. This section classifies the methods used into 3 classes - non-neural network methods, methods using 2D CNNs and methods using 3D CNNS. Methods in each of these classes are described in the following subsections.

\subsection{Non-neural Network Methods}
\label{rel:1}
Decision Forests played a major role in hand pose estimation from depth images in earlier methods \cite{Ref7, Ref17, Ref19, Ref23, Ref24, Ref25, Ref31}. These methods required manual feature selection and performed significantly lower than more recent methods. Approaches \cite{Ref8, Ref15} based on particle swarm optimization (PSO) require an increased compute because of the large number of particles involved. Tagliasacchi et al. \cite{Ref20} use iterative closest point (ICP) approach and Qian et al. \cite{Ref21} use a combination of ICP and PSO.

\subsection{2D Convolutional Neural Networks}
\label{rel:2}
A majority of methods \cite{Ref3, Ref4, Ref5, Ref11, Ref13, Ref16, Ref27, Ref28, Ref38, Ref40, Ref41} involve the use of 2D CNNs. Guo et al. \cite{Ref3, Ref4} partition the convolution feature maps into regions and integrate the results from multiple regressors on each region. Chen et al. \cite{Ref5} improve upon this method by using a guided initial hand pose. Ge et al. \cite{Ref11} project the 2D depth map onto three orthogonal planes and fuse the result of 2D CNNs trained on each plane. Tompson et al. \cite{Ref13} were first to predict heatmaps representing 2D joint positions in the depth image. Madadi et al. \cite{Ref16} use a hierarchical tree-structured CNN for estimation. Oberwerger et al. \cite{Ref28} bettered their work \cite{Ref27} by introducing data augmentations, improved localization and a newer network architecture. Sinha et al. \cite{Ref38} use a matrix completion method. Yang et al. \cite{Ref40} use a CNN to classify depth images into different types and later perform regression on them. Xu et al. \cite{Ref41} apply Lie group theory to estimate poses.

All of the above methods use 2D filters in 2D CNNs to extract features from a depth map. As a result of this 2D mapping, important spatial information is lost while predicting 3D joint locations. Our method avoids this loss of information by employing a 3D CNN which can efficiently capture spatial information to map 3D joint locations from depth images.

\subsection{3D Convolutional Neural Networks}
\label{rel:3}
Volumetric representations of depth maps in the form of binary variables representing a voxel grid were first proposed by Wu et al. \cite{Ref33}. They used a convolutional deep belief network to map the probability distribution for each binary variable to represent the 3D geometric shape. Inspired from Maturana et al. \cite{Ref36}, where they show different representations for representing a voxel grid, we use a binary voxel grid to represent the depth map input in our model.

The most recent methods \cite{Ref6, Ref10, Ref12} make use of 3D CNNs for joint estimations. Ge et al. \cite{Ref10} and Deng et al. \cite{Ref12} make use of truncated signed distance function (TSDF) to represent the depth map points in a volumetric shape. Moon et al. \cite{Ref6} use input similar to our method but output a heatmap showing per-voxel likelihood of each joint. The time and compute taken in their method is significantly reduced in our method as shown in Table \ref{tab:2}.

\section{Input to the Network}
\label{input}
This section discusses the steps involved in generating the input which is fed to our model. We localize the hand from its surroundings, voxelize it to capture the 3D spatial information and perform data augmentations to generalize our model.

\subsection{Hand Localization}
\label{in:1}
\begin{figure}[b]
\sidecaption
% For example, with the graphicx style use
\includegraphics[scale=.5]{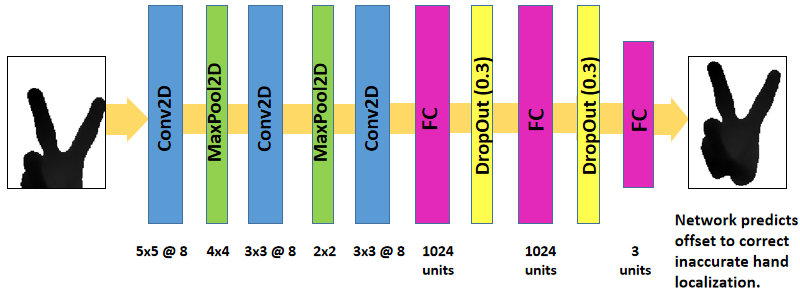}
%
% If no graphics program available, insert a blank space i.e. use
%\picplace{5cm}{2cm} % Give the correct figure height and width in cm
%
\caption{ Hand localization network. An offset to correct an inaccurate hand localization is predicted by the network from the previous reference point to the new reference point.}
\label{fig:2}       % Give a unique label
\end{figure}

Segmenting the hand from its surrounding background is a prerequisite to hand pose estimation. Hand localization feeds the network the complete hand discarding much of the unnecessary background information that comes along with it when using our method on real-time images. Inspired from Obwerger et al. \cite{Ref27, Ref28} we segment the hand using depth-thresholding and calculate its center of mass. A 3D bounding box is built around the center of mass. We then train a 2D CNN, as shown in Figure \ref{fig:2} to refine the hand locations by regressing one reference point per frame based on the center of mass calculated for that frame. We use data augmentations to generalize the input as described in Section \ref{in:3}.

\subsection{Volumetric Representation}
\label{in:2}
The images captured by depth cameras are a function of the 2D depth image pixel coordinates $(p,q)$ represented as $D(p,q)$. We project these pixel coordinates $(p,q)$ to world coordinates $(x, y, z)$ using the camera intrinsic parameters $f_p$, $f_q$.

\begin{equation}
(x, y, z) = \left( \frac{p}{f_p}D(p,q), \frac{q}{f_q}D(p,q), D(p,q) \right)
\end{equation}

We discretize the 3D projected points into a voxel grid based on a predefined voxel size of $10 mm^3$ . Each voxel $V[x,y,z]=1$ if it is occupied by a depth point and is set to $0$ if it is not. Using the reference point obtained in Section \ref{in:1}, we set the voxel grid to a size of $G^3$ voxels where $G=200$ to cover the entire camera coordinate system around the reference point. A similar method to voxelize has also been used in \cite{Ref9}.

\subsection{Data Augmentation}
\label{in:3}
In order to make our model robust to different hand
sizes and global orientations we perform data augmentations on the 3D projected depth map points. We perform scaling by a random factor in the range [0.7, 1.2] to resize the input to a random size. We then perform translation on the voxel grid by a random number of voxels in the range [-7, 7] linearly. Finally, we perform rotation of the voxel grid in the XY plane by an angle chosen randomly from the range [-40, 40] degrees. For a voxel in the 3D space $v$ , we perform the following operations as described above to change it into a voxel $v''$ :

\begin{equation}
v' = v.s + t
\end{equation}
\begin{eqnarray}
\begin{avm}
\[ $v_x''$ \\
$v_y''$ \]
\end{avm}
 = 
\begin{avm}
\[ $\cos\theta$ & $-\sin\theta$ \\
   $\sin\theta$ & $\sin\theta$ \]
\end{avm}
.
\begin{avm}
\[$v_x'$ \\
$v_y'$\]
\end{avm} ;v_z'' = v_z'
\end{eqnarray}

Here $s$ is the scale factor and $t$ is the number of voxels by which translation is performed. $v'$ is the voxel after performing translation and scaling with $v_x, v_y, v_z$ being its components along X, Y, Z axes respectively. We then perform rotation in the XY
plane by the angle $\theta$ resulting in the augmented voxel $v''$.
The entire training set is generated by augmenting the data in the original training set.

\begin{figure}[t]
\sidecaption[t]
% For example, with the graphicx style use
\includegraphics[scale=.55]{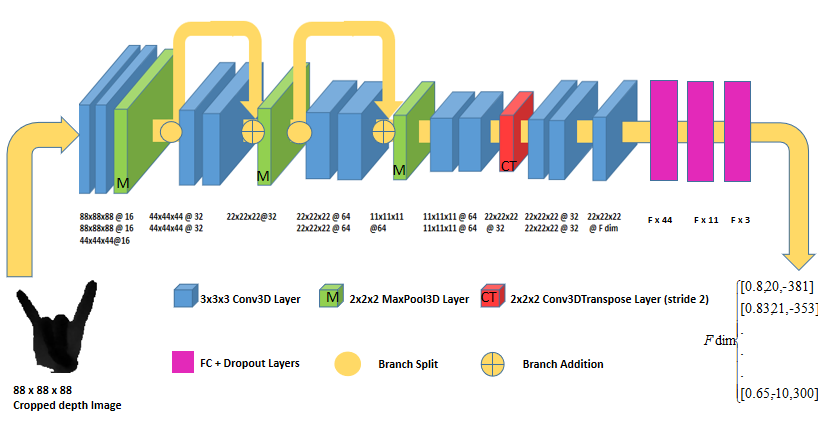}
%
% If no graphics program available, insert a blank space i.e. use
%\picplace{5cm}{2cm} % Give the correct figure height and width in cm
%
\caption{Overall architecture of FastV2C-HandNet. Each Conv3D layer comprises of 3D Convolutional Layer (3x3x3) + Batch Normalization + ReLU Activation. Each MaxPool3D layer consists of a 3D Max Pooling layer with size (2x2x2). The Conv3DTranspose layer has kernel size of (2x2x2) and stride 2.$F$ are the number of joints per hand pose. The network takes in a voxelized grid input and regresses 3D world joint coordinates.}
\label{fig:3}       % Give a unique label
\end{figure}

\section{FastV2C-HandNet (Our Method)}
\label{method}
In this section we describe the network architecture used along with the implementation details for training our model. The input passes through a series of 3D convolutional blocks with two residual connections \cite{Ref26} in between to facilitate increased depth at a reduced complexity.

\subsection{Network Architecture}
\label{method: 1}
As seen in Figure \ref{fig:3}, the input to our model is an 88x88x88 voxelized depth image. Our proposed model is composed of eleven 3D convolutional layers. We use three 3D max pooling layers to down-sample the input size so that the most important features are captured. Two residual connections \cite{Ref26} are adopted after the first and the second max pooling layers to increase the dimension size. The filter size of each pair of Conv3D layers increases from 16 to 64 with the last Conv3D layer having a filter size of $F$ , which is the number of joints in the hand pose.To up-sample the input we then introduce a 3D transpose convolutional layer with a filter size of 32 and a stride of 2. The output from this layer passes through a pair of convolutional layers with 32 filters. All
the Conv3D layers and the Conv3DTranspose layer are followed by Batch Normalization and Rectified Linear Unit (ReLU) activation. The Conv3D layers each have a kernel size of 3 except the last layer which has a kernel size of 1. The MaxPool3D layers have a stride of 2 each. The Conv3DTranspose layer has a filter size of 32 with kernel size 2 and stride 2. This completes the feature extraction part of our network.

The features extracted through the above network are then passed through three fully connected (FC) layers to regress the joint coordinates. For $F$ number of joints in a pose each fully connected layer has $F$ x $c$ units, where c = 44, 11 and 3 respectively. Each of the first two fully connected layers is followed by a DropOut \cite{Ref42} layer with a dropout rate of 0.5 to prevent the network from overfitting on the training set. We get a $F$ x $c$ dimensional output vector with each row corresponding to the 3D world coordinates for each joint.
\subsection{Implementation}
\label{method: 2}
We segment the hand from the depth image and
voxelize the depth map, on which we perform data augmentations as described in Sections \ref{in:1}, \ref{in:2} and \ref{in:3}. We input an 88x88x88 voxel grid obtained by cropping the 96x96x96 3D projection. We use the Mean Squared Error (MSE) loss function $L$ to calculate the loss between the estimated joint locations and the ground truth joint locations.

\begin{equation}
L = \frac{1}{N} \sum_{n=1}^{N} \sum_{i,j,k} (i-i_n)^2 + (j-j_n)^2 + (k-k_n)^2
\end{equation}

Here $N$ are the number of joints per frame and $i_n, j_n, k_n$ are the ground truth joint locations of the $n^{th}$ joint.

The weights are updated using the Adam Optimizer \cite{Ref14} with a learning rate of $3.0$x$10^{-4}$ . We use a mini-batch size of 4 and train the network for 3 epochs. The kernel weights are initialized from a zero-mean normal distribution with
$\sigma = 0.005$ and the biases are initialized with zeros. To demonstrate the speed of our network we trained it on a single NVIDIA Tesla P100 GPU and got state-of-the-art-results in terms of speed. (Section \ref{exp:3})

\begin{figure}
\sidecaption
% For example, with the graphicx style use
\includegraphics[scale=.25]{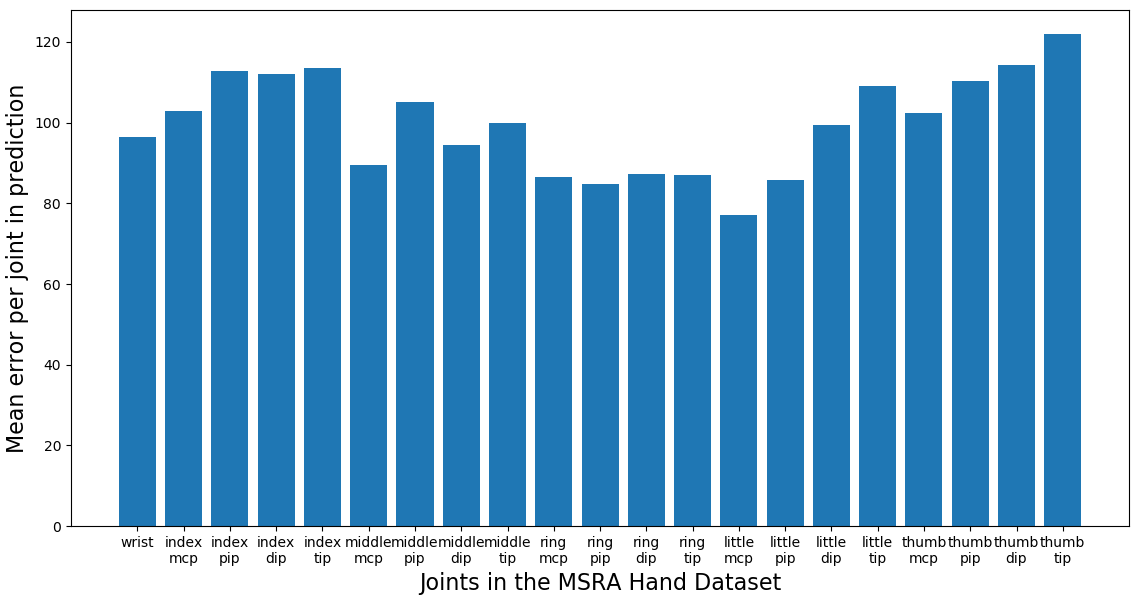}
%
% If no graphics program available, insert a blank space i.e. use
%\picplace{5cm}{2cm} % Give the correct figure height and width in cm
%
\caption{Mean error per joint for the 21 joints in the MSRA Hand Dataset.}
\label{fig:4}       % Give a unique label
\end{figure}

\begin{figure}
\sidecaption
% For example, with the graphicx style use
\includegraphics[scale=.25]{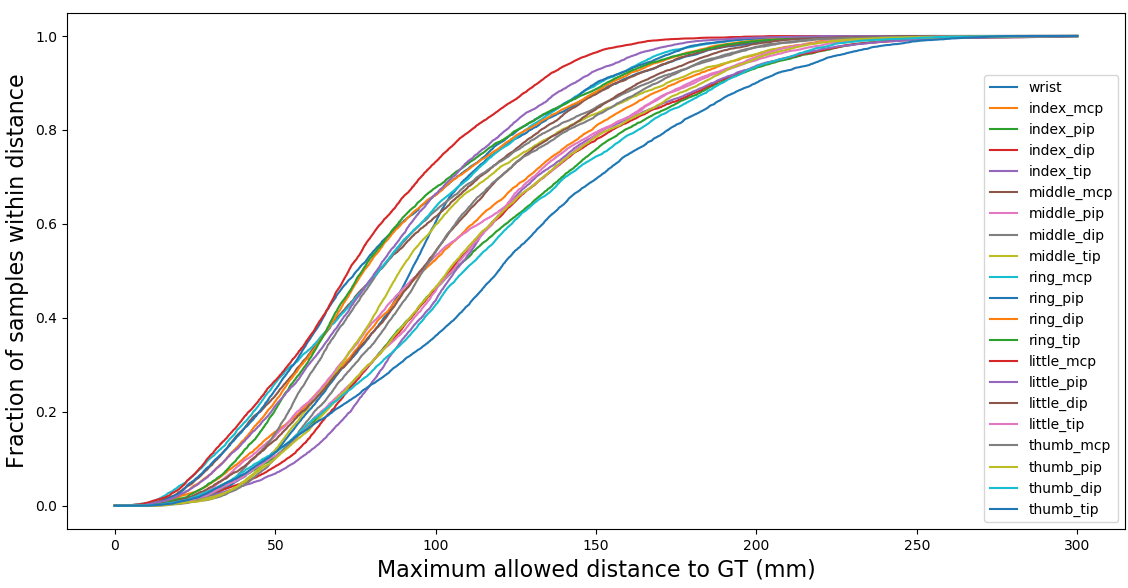}
%
% If no graphics program available, insert a blank space i.e. use
%\picplace{5cm}{2cm} % Give the correct figure height and width in cm
%
\caption{Fraction of success frames vs maximum distance from ground truth location in experiments performed on the MSRA Hand Dataset.}
\label{fig:5}       % Give a unique label
\end{figure}

\section{Experiments}
\label{exp}
We perform extensive experiments and provide results in this section. The experiments are performed using a publicly available benchmark dataset. The evaluation metrics are then discussed which are followed by the results of our experiments.

\subsection{Dataset}
\label{exp:1}
We evaluate our model on the MSRA Hand Pose Dataset \cite{Ref19} which is a publicly available benchmark dataset. The Dataset is composed of depth images captured from 9 different subjects depicting 17 hand pose gestures each. Each gesture has about 500 frames. The creators use Intel\textquotesingle s Creative Interactive Gesture Camera \cite{Ref32} to obtain more than 76K depth images with 21 hand joints per image. Since there are no explicitly defined train and test sets, we train on 8 subjects and test on the remaining one subject. This is repeated 9 times for 9 different test subjects.

\subsection{Evaluation Metrics}
\label{exp:2}
We use 2 evaluation metrics to evaluate the performance of our model:
\begin{itemize}
\item We use the 3D mean joint error to evaluate the accuracy of our model. This metric gives the average error per joint in all the samples.
\item As our second metric, we plot the fraction of frames having all predicted joints\textquotesingle Euclidean distance from the ground truth less than a maximum value \cite{Ref34}.
\end{itemize}

\subsection{Results}
\label{exp:3}
We train our network on the MSRA Hand Dataset \cite{Ref19} as described in Section \ref{exp:1}. We plot the mean error per joint as shown in Figure \ref{fig:4}. We also plot the fraction of samples having distance less than a maximum from the ground truth. As seen in Figure \ref{fig:5}, predicted joints in more than 50 percent of the samples fed to the network have less than 100mm distance from the ground truth joints. From Figure \ref{fig:4} we find that the lowest error across all samples is recorded for little\_icp joint and the highest error for thumb\_tip which can also be intuitively understood from Figure \ref{fig:5}. The total training time taken by our network is about 7 hours on the MSRA Hand Dataset which to the best of our knowledge, is significantly less compared to all previous hand pose estimation methods using 3D CNNs.

\section{Comparison with state-of-the-arts}
\label{comp}
We compare the running time of our method with
state-of-the art methods \cite{Ref6, Ref10, Ref12} on the MSRA Hand Dataset. As shown in Table \ref{tab:1}, our method significantly reduces the time complexity thereby achieving faster results than any of the previous methods. Our network is end-to-end which avoids the time-consuming multistage pipelines \cite{Ref7, Ref8, Ref15} and feedback loops \cite{Ref43} used in previous methods.

We also list a separate comparison with V2V-PoseNet \cite{Ref6} which gives the lowest error currently for this task to show that our method reduces the training and prediction time taken significantly while retaining the 3D spatial information of the depth image. This is shown in Table \ref{tab:2}.

\begin{table}
\caption{Table 1: Comparison of the proposed method (FastV2C-HandNet) with state-of-the-art methods based on time taken per frame for joint prediction. Our method outperforms all previous methods.}
\label{tab:1}       % Give a unique label
\begin{tabular}{p{6cm}p{3cm}}
\hline\noalign{\smallskip}
\textbf{Methods} & \textbf{Time per frame(ms)}  \\
\noalign{\smallskip}\hline\noalign{\smallskip}
DeepPrior++ \cite{Ref28} & 33.33 \\
V2VPoseNet \cite{Ref6} & 28.5 \\
PointNet \cite{Ref45} & 23.92\\
HandPointNet \cite{Ref46} & 20.833 \\
Madadi \textit{et al.} \cite{Ref16} & 20.0 \\
CascadedPointNet \cite{Ref47} & 14.3 \\
CrossingNets \cite{Ref44} & 11 \\
Ge \textit{et al.} \cite{Ref12} & 7.9 \\
Latent2.5D \cite{Ref48} & 6.89 \\
REN \cite{Ref4} & 0.31 \\
\textbf{FastV2C-HandNet (Ours)} & \textbf{0.185} \\
\noalign{\smallskip}\hline
\end{tabular}
\end{table}

\begin{table}
\caption{Table 2: Comparison of the proposed method’s (FastV2C-HandNet) total training time and model size on the MSRA Hand Dataset \cite{Ref19} with V2V-PoseNet \cite{Ref6}.}
\label{tab:2}       % Give a unique label
\begin{tabular}{p{5cm}p{3cm}p{2cm}}
\hline\noalign{\smallskip}
\textbf{Methods} & \textbf{Total Training Time} & \textbf{Model Size}  \\
\noalign{\smallskip}\hline\noalign{\smallskip}
V2V-PoseNet \cite{Ref6} & 12 hours & 457 MB \\
\textbf{FastV2C-HandNet (Ours)} & \textbf{7 hours} & \textbf{42 MB} \\
\noalign{\smallskip}\hline
\end{tabular}
\end{table}

\section{Conclusion}
\label{conc}
In this paper, we presented a novel voxel-to-coordinate model for hand pose estimation. We segment the hand from its background and localize it by regressing reference points using a 2D CNN network. To overcome the drawbacks in methods using 2D CNNs and random forests we voxelize the depth map to preserve the spatial information. In order to make our network robust to real-world scenarios we perform data augmentations on our training set before feeding it to the network. The network is trained in an end-to-end manner which significantly reduces the time and compute power required. We evaluate our model on one publicly available hand dataset. Through experiments we show that our model performs faster than the state-of-the-art methods for hand pose estimation which makes it more suitable to be used in real-world scenarios.

A limitation of our method is a slight decrease in accuracy with the increase in speed. Due to the regression of the hand joints directly as 3D coordinates instead of a per-voxel likelihood heatmap as in \cite{Ref6}, a decrease in accuracy is observed. An extension of this work will present a modification of this algorithm that can maintain the speeds achieved by this method without affecting the accuracy. This forms a scope for future research.

%%%%%%%%%%%%%%%%%%%%%%%% referenc.tex %%%%%%%%%%%%%%%%%%%%%%%%%%%%%%
% sample references
% %
% Use this file as a template for your own input.
%
%%%%%%%%%%%%%%%%%%%%%%%% Springer-Verlag %%%%%%%%%%%%%%%%%%%%%%%%%%
%
% BibTeX users please use
% \bibliographystyle{}
% \bibliography{}
%

\end{document}